\def\year{2022}\relax
\documentclass[letterpaper]{article} 
\usepackage{aaai22}  
\usepackage{times}  
\usepackage{helvet}  
\usepackage{courier}  
\usepackage[hyphens]{url}  
\usepackage{graphicx} 
\usepackage{natbib}  
\usepackage{caption} 
\DeclareCaptionStyle{ruled}{labelfont=normalfont,labelsep=colon,strut=off} 
\title{Virtual Replay Cache}
\author {
    Brett Daley,\ 
    Christopher Amato
}
\affiliations {
    Khoury College of Computer Sciences, Northeastern University, Boston, MA 02115\\
    daley.br@northeastern.edu, c.amato@northeastern.edu
}

\nocopyright

\usepackage{import}
\usepackage{paper_header}

\begin{document}

\maketitle

\begin{abstract}
    Return caching is a recent strategy that enables efficient minibatch training with multistep estimators (e.g.\ the $\lambda$-return) for deep reinforcement learning.
    By precomputing return estimates in sequential batches and then storing the results in an auxiliary data structure for later sampling, the average computation spent per estimate can be greatly reduced.
    Still, the efficiency of return caching could be improved, particularly with regard to its large memory usage and repetitive data copies.
    We propose a new data structure, the Virtual Replay Cache (VRC), to address these shortcomings.
    When learning to play Atari 2600 games, the VRC nearly eliminates DQN($\lambda$)'s cache memory footprint and slightly reduces the total training time on our hardware.
\end{abstract}

\section{Introduction}

The $\lambda$-return is a popular multistep estimator for credit assignment technique in reinforcement learning \cite{sutton1998reinforcement}.
Estimates are constructed by combining all future $n$-step returns in an exponentially weighted average \cite{watkins1989learning}.
The $\lambda$-return has a number of appealing properties:
an implicit recency heuristic, a mechanism for balancing the bias-variance tradeoff \cite{kearns2000bias}, and an efficient recursive structure.

Modern reinforcement learning has increasingly turned to complex function approximators (such as deep neural networks) to learn in high-dimensional environments.
The nonlinearity of these approximators generally renders incremental temporal-difference learning and eligibility traces ineffective, which has encouraged the adoption of offline experience replay \cite{lin1992self} as an alternative training mode.
Combining $\lambda$-returns with experience replay is expensive since each return must be computed individually, each of which in turn requires multiple value estimates (i.e.\ multiple costly function approximator evaluations).

\citet{daley2019reconciling} proposed \textit{return caching} as part of the DQN($\lambda$) framework to mitigate this expense.
The $\lambda$-returns for a subset of the replay memory are efficiently precomputed in sequential batches (``blocks'') and then stored in an auxiliary data structure (``cache'').
The cache then assumes the typical role of the replay memory for minibatch sampling, with the added benefit of replacing the target network of Deep Q-Network (DQN) \cite{mnih2015human}.
Due to the recursive nature of the $\lambda$-return, return caching significantly reduces the average computation spent per estimate---only one value estimate is needed per $\lambda$-return.
Unfortunately, the cache consumes additional memory (particularly when states or actions are high dimensional) and requires the extraneous movement of data.

Our contribution is the \textit{Virtual Replay Cache} (VRC).
Rather than storing state-action pairs directly in the cache, their locations in the replay memory are tracked and then dynamically dereferenced as needed.
This avoids unnecessary data copies and saves significant memory.
When compared to DQN($\lambda$), the VRC almost eliminates cache memory consumption and slightly reduces runtime.
Our open-source implementations of the VRC and DQN($\lambda$) are publicly available.\footnote{
    \url{https://github.com/brett-daley/virtual-replay-cache}
}

\begin{figure*}[t]
    \centering
    \includegraphics[width=\textwidth]{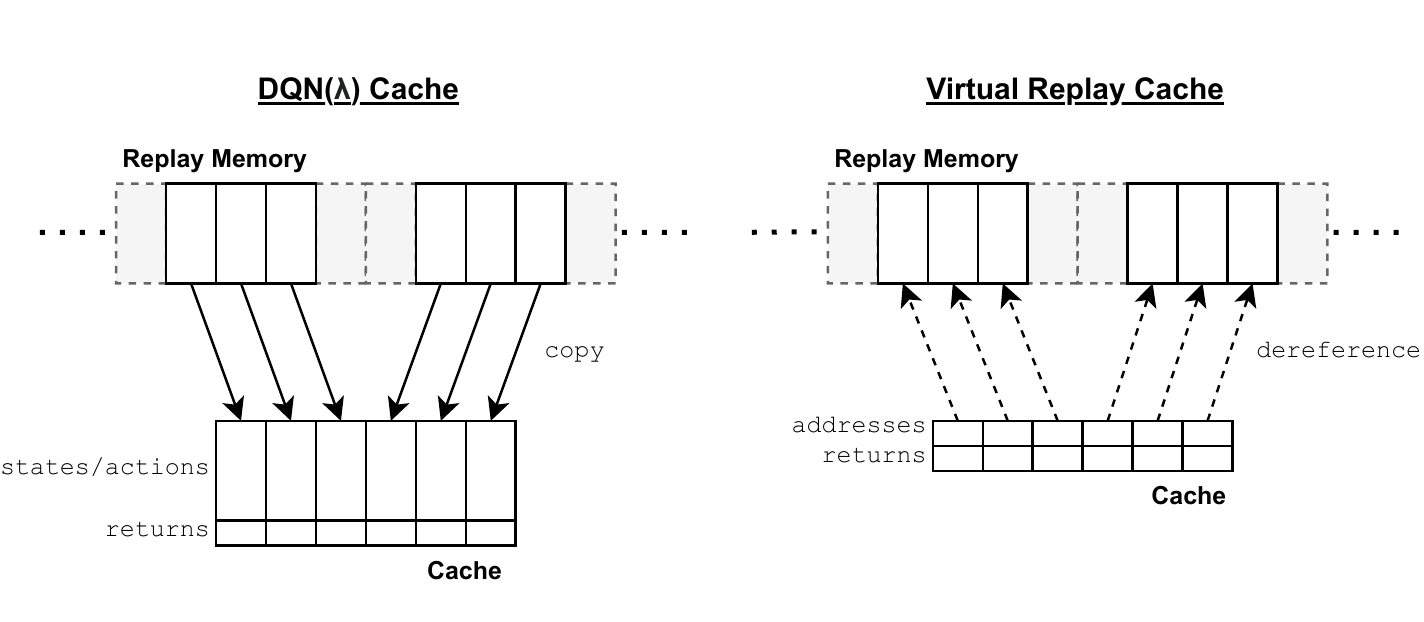}
    \caption{
        Graphical comparison of the DQN($\lambda$) cache (left) and the VRC (right).
        The DQN($\lambda$) cache wastefully duplicates state-action pairs that already exist in the replay memory---a costly operation for high-dimensional MDPs.
        Instead, the VRC stores only the \textit{locations} of the state-action pairs, referencing them indirectly as needed and saving substantial memory.
    }
    \label{fig:illustration}
\end{figure*}

\section{Background}

Let
$M \coloneqq (\mathcal{S}, \mathcal{A}, P, R, \gamma)$
be a Markov Decision Process (MDP) with finite sets of states $\mathcal{S}$ and actions $\mathcal{A}$, transition probabilities $P(s_{t+1}|s_t,a_t)$, reward function ${r_t \coloneqq R(s_t, a_t)}$, and discount factor $\gamma \in [0,1]$.
Given a value function
$V \colon \mathcal{S} \mapsto \mathbb{R}$
and a policy
$\pi \colon \mathcal{S} \mapsto \mathcal{A}$,
a fundamental question in reinforcement is how to update $V(s_t)$ based on sampled experiences such that it more closely approximates
$V^\pi(s_t) \coloneqq \mathbb{E}_\pi [\sum_{k=0}^\infty \gamma^k r_{t+k}]$.
This is typically accomplished by constructing a target $G_t$ and moving the current estimate $V(s_t)$ towards it.
The simplest target is the $n$-step return:
\begin{equation}
    G^{(n)}_t \coloneqq \sum_{k=0}^{n-1} \gamma^i r_{t+k} + \gamma^n V(s_{t+n})
    .
\end{equation}
To reduce variance, any convex combination of $n$-step returns can be substituted as the target \cite{sutton1998reinforcement}.
The $\lambda$-return is one such possibility, defined as
\begin{equation}
    \label{eq:lambda_return_truncated}
    \Lambda^{(N)}_t \coloneqq (1 - \lambda) \sum_{n=1}^{N-1} \lambda^{n-1} G^{(n)}_t + \lambda^{N-1} G^{(N)}_t
\end{equation}
for a trajectory starting at timestep $t$ and terminating at timestep $t + N$.
(The formula also generalizes to non-terminal trajectories by letting $N \to \infty$ with $\gamma < 1$.)
The $\lambda$-return is recursive:
\begin{equation}
    \label{eq:lambda_return_recursive}
    \Lambda^{(N)}_t = G^{(1)}_t + \gamma \lambda \left( \Lambda^{(N-1)}_{t+1} - V(s_{t+1}) \right)
    .
\end{equation}
This property underpins the online implementation of eligibility traces \cite{klopf1972brain, barto1983neuronlike, sutton1984temporal}.
It also implies that an offline sequence of $\lambda$-returns can be computed efficiently in reverse-chronological order \cite{daley2019reconciling}.
The DQN($\lambda$) algorithm leverages this by efficiently computing $\lambda$-returns in $B$-length blocks sampled randomly from $\mathcal{D}$ and then storing them in a cache $\mathbb{C}$ of size $S$.
The cache is refreshed every $C$ timesteps using the latest network parameters $\theta$, thereby replacing DQN's target network.

For the purposes of model-free control, DQN($\lambda$) learns a Q-function
$Q \colon \mathcal{S} \times \mathcal{A} \mapsto \mathbb{R}$
instead of the state-value function $V$.
The expression $V(s_t) = \max_{a \in \mathcal{A}} Q(s_t, a)$ is therefore substituted into Equation (\ref{eq:lambda_return_truncated}).
This makes the $\lambda$-return formulation equivalent to Peng's Q($\lambda$) \cite{peng1994incremental}, which is a biased estimator but still converges to $Q^*$ under certain technical conditions \cite{kozuno2021revisiting}.

Training the $Q$-function is achieved by minimizing a mean squared error loss
\begin{equation}
    L(\theta) \coloneqq \mathbb{E} \! \left[ (\Lambda - Q(s,a;\theta))^2 \right]
\end{equation}
where the expectation is taken over experiences $(s, a, \Lambda)$ sampled uniformly from $\mathbb{C}$.
Following the DQN algorithm, one minibatch of training is conducted every $F$ timesteps.

\section{Virtual Cache Storage}
\label{sect:virtual_storage}

A disadvantage of DQN($\lambda$) is its high memory consumption since it duplicates state-actions pairs that are promoted into the cache $\mathbb{C}$ from the replay memory $\mathcal{D}$.
This is expensive when states or actions are high dimensional.

Another potential concern is that the process of copying data is slow with respect to wall-clock time.
Although a good DQN($\lambda$) implementation will pre-allocate memory for the cache (thereby avoiding the repeated expense of memory allocation), the cost of transferring data from the replay memory to the cache is still likely to be non-negligible.
Avoiding this extraneous movement of data could reduce runtime while also simplifying the implementation.

One key observation is that the storage of state-action pairs in DQN($\lambda$)'s cache serves a purely logical function:
indication of which state-action pairs are currently eligible for minibatch sampling.
Since these data already exist in the replay memory, it is redundant and inefficient to explicitly copy them into a separate data structure when they could have been retrieved from the original source just as easily.
We can thus apply the principle of \textit{indirection};
instead of storing data in a physical cache as DQN($\lambda$) does, we need only some indicator of which data currently reside ``in'' the cache, as well as their locations in the replay memory.

If the replay memory is implemented as an array, then cache membership can be easily tracked by storing the numerical index of each experience.
This is related to the idea behind Stratified Experience Replay \cite{daley2021stratified}, where indices (rather than data) are manipulated for the purposes of efficiently implementing a non-uniform sampling distribution.
Here, we take a similar approach, but with the goal of saving memory when implementing cache-based learning.
As a result, no experiences need to be copied at any point during the cache building process;
their indices are recorded (along with their pre-computed $\lambda$-returns) and then used to locate the state-action pairs later as needed.
We illustrate this concept in Figure~\ref{fig:illustration}.
When the memory footprint of a state-action pair is much larger than that of a single integer, then the VRC saves a considerable amount of memory.

Our indirect addressing technique is precisely why we call our method the \textit{Virtual} Replay Cache.
We present pseudocode for training DQN with the VRC in Algorithm~\ref{algo:vrc}.
Most importantly, in conjunction with the aforementioned computational benefits, the VRC is a pure refactor of DQN($\lambda$);
the empirical learning performances of the two methods are identical.

\section{Experiments}
\label{sect:experiments}

We evaluate the VRC's computational performance in a number of emulated Atari 2600 games \cite{bellemare2013arcade, brockman2016openai}.
Our experimental setup is identical to that of \citet{daley2019reconciling}, which in turn closely follows the DQN procedures from \citet{mnih2015human}.
We compare the VRC against DQN($\lambda$), testing both methods with cache size $S=80000$ and block size $B=100$ (see Table~\ref{table:hyperparameters} in the appendix for all hyperparameter values).
We train the agents for 5 million timesteps (20 million game frames) and average all results over five trials, with standard deviation reported.

\begin{table}[t]
    \centering
    \caption{
        Average runtime in hours for DQN($\lambda$) and the VRC.
        Standard deviation is reported.
        The parentheses contain the normalized mean runtimes:
        $100 \times \text{VRC} \mathbin{/} \text{DQN}(\lambda)$.
    }
    \label{table:runtime}
    \begin{tabular}{lcc}
        \toprule
         & DQN($\lambda$) & VRC\\
        \midrule
        Beam Rider     & $2.633 \pm 0.005$ & $2.600 \pm 0.003$ ($98.8\%$)\\
        Breakout       & $2.603 \pm 0.004$ & $2.575 \pm 0.012$ ($98.9\%$)\\
        Pong           & $2.551 \pm 0.011$ & $2.524 \pm 0.019$ ($99.0\%$)\\
        Q*bert         & $2.611 \pm 0.015$ & $2.579 \pm 0.013$ ($98.8\%$)\\
        Seaquest       & $2.591 \pm 0.004$ & $2.570 \pm 0.015$ ($99.2\%$)\\
        Space Invaders & $2.551 \pm 0.016$ & $2.519 \pm 0.012$ ($98.7\%$)\\
        \bottomrule
    \end{tabular}
\end{table}

\begin{table}[t]
    \centering
    \caption{
        Cache memory usage for DQN($\lambda$) and the VRC, assuming $S=80000$, when playing Atari 2600 games with the preprocessing procedures from \citet{mnih2015human}.
        Units are bytes (B) or megabytes (MB):
        $1\ \text{MB} = 2^{20}\ \text{B}$.
        The VRC's size constitutes only $0.028\%$ of the DQN($\lambda$) cache's memory footprint:
        a reduction of over $99.9\%$.
    }
    \label{table:ram_usage}
    \begin{tabular}{lrr}
        \toprule
         & DQN($\lambda$) & VRC\\
        \midrule
        Per Experience & 28229 B & 8 B\\
        Cache          & 2154 MB & 0.61 MB\\
        \bottomrule
    \end{tabular}
\end{table}

\paragraph{Runtime}
We evaluate the total runtime of both algorithms (Table~\ref{table:runtime}).
Our test hardware consists of an Intel i7-7700K CPU and an NVIDIA GeForce GTX 1080 GPU.
We measure an approximately 1\% runtime reduction (about 1.5 minutes) for the VRC compared to DQN($\lambda$) on all six games.
This confirms our hypothesis that the excessive copies made by DQN($\lambda$) negatively impact its runtime;
however, even though the speedup is statistically significant, it is unlikely to offer much practical benefit on its own.

\paragraph{Memory Usage}
We compute the per-experience and total cache memory usage for both DQN($\lambda$) and the VRC in Table~\ref{table:ram_usage};
see Appendix~\ref{app:memory_usage} for a breakdown of these values.
The VRC requires only 8 bytes of memory per experience in the cache, reducing the total cache memory usage of DQN($\lambda$) by over 99.9\% (about 2.5 gigabytes).
As such, the VRC almost entirely erases the memory footprint of cache-based training.
Practically speaking, the memory savings are small compared to the total size of DQN's replay memory, but they demonstrate the favorable scalability of the VRC;
these savings scale linearly with respect to both the dimensionality of the state-action pairs and the chosen cache size $S$, meaning greater benefits could be seen in more-complex environments than Atari 2600 games.

\section{Conclusion}

We introduced the Virtual Replay Cache (VRC), a data structure for efficiently training deep reinforcement learning agents on minibatches of replayed $\lambda$-returns (or other multistep estimators).
Through its novel indirect addressing scheme, the VRC occupies a much smaller memory footprint and executes slightly faster than DQN($\lambda$).
Most significantly, the VRC is a modular structure that can be straightforwardly extended to other learning algorithms beyond the DQN agent considered here in our work.

\begin{algorithm}[t]
    \caption{DQN($\lambda$) with Virtual Replay Cache}
    \label{algo:vrc}
    \begin{algorithmic}
        \State Prepopulate replay memory $\mathcal{D}$ with $K$ experiences
        \State Initialize parameter vector $\theta$ randomly
        \State Initialize environment state $s_1$
        \For {$t = 1, 2, \dots$\ until convergence}
            \If{$t \bmod C = 1$}
                \State \Call{train}{$\mathcal{D}$}
            \EndIf
            \State Execute action $a_t \sim \pi(\cdot \mid s_t; \theta)$, receive reward $r_t$,
            \State \quad observe next state $s_{t+1}$
            \State Store experience $(s_t, a_t, r_t)$ in $\mathcal{D}$
        \EndFor
        \State
        \Function{train}{$\mathcal{D}$}
            \State $\mathbb{C} \gets $ \Call{build-cache}{$\mathcal{D}$}
            \For{$C \mathbin{/} F$ iterations}
                \State Sample minibatch $(j, \Lambda_j)$ randomly from $\mathbb{C}$
                \State Retrieve minibatch $(s_j, a_j, r_j)$ from $\mathcal{D}$
                \State Minimize $\smash{(\Lambda_j - Q(s_j, a_j; \theta))^2}$ w.r.t.\ $\theta$
            \EndFor
        \EndFunction
        \State
        \Function{build-cache}{$\mathcal{D}$}
            \State $\mathbb{C} \gets \{\}$
            \Repeat
                \State Sample block $\{(s_t, a_t, r_t)\}_{t=k}^{k+B+1}$ from $\mathcal{D}$
                \State Compute $\lambda$-returns for $t = k, \dots, k+B$:
                \State \quad $v \gets \max\limits_{a' \in \mathcal{A}} Q(s_{t+1},a';\theta)$ \enskip ($0$ if terminal $s_{t+1}$)
                \State \quad $\Lambda_t \gets r_t + \gamma (\lambda \Lambda_{t+1} + (1-\lambda) v)$
                \State Store results in cache: $\mathbb{C} \gets \mathbb{C} \cup \{(t, \Lambda_t)\}_{t=k}^{k+B}$
            \Until{$|\mathbb{C}| = S$}
            \State \Return $\mathbb{C}$
        \EndFunction
    \end{algorithmic}
\end{algorithm}

\clearpage

\bibliography{references.bib}

\appendix

\section{Memory Usage Calculation}
\label{app:memory_usage}

In this section, we describe how the memory usage values in Table~\ref{table:ram_usage} were calculated.
The values could differ depending on the implementation, but we assumed the most-favorable conditions for both algorithms by choosing the smallest feasible representations from the standard data types.

\paragraph{DQN($\lambda$)}
Each experience in the DQN($\lambda$) cache comprises a state, an action, and a $\lambda$-return.
A state corresponds to four single-channel grayscale images stacked together, each with pixel dimensions $84 \times 84$.
These images can be stored as 8-bit (1-byte) unsigned integers, totaling $84 \times 84 \times 4 = 28224$ bytes per image.
The discrete Atari actions can also be represented as 1-byte integers (there are fewer than 256 actions for each game).
We assume that single-precision (32-bit, 4-byte) floating point numbers are sufficient for the $\lambda$-return calculations.
This brings the total to 28229 bytes per experience in the cache.
Notice that the vast majority of the cache's memory consumption is due to the storage of the environment states, particularly because the agent must learn from high-dimensional sensory data in the Atari 2600 domain.

\paragraph{VRC}
Each experience in the VRC comprises only an array address (that points to the state-action pair in the replay memory) and its corresponding $\lambda$-return.
We assume that the address is represented as a 32-bit (4-byte) integer, which should be capable of indexing any location in a reasonably sized replay memory (up to about 4 billion transitions, 4000 times larger than the capacity in our experiments).
As before, we assume that the $\lambda$-return is a 32-bit (4-byte) floating point number.
We therefore arrive at a mere total of 8 bytes per experience in the VRC;
by avoiding duplicated state-action pairs in the cache, the VRC nearly erases the additional memory requirement of cache-based training.

\begin{table}[t]
    \renewcommand{\arraystretch}{1.5}
    \small
    \centering
    \caption{Hyperparameters for DQN($\lambda$) and the VRC.}
    \begin{tabular}{l c c}
        \toprule
        \textbf{Hyperparameter} & \textbf{Symbol} & \textbf{Value} \\
        \midrule
        minibatch size               &          & 32\\
        replay memory capacity       &          & 1000000 \\
        cache refresh frequency      & $C$      & 10000\\
        effective training frequency & $F$      & 4\\
        discount factor              & $\gamma$ & 0.99\\
        replay memory prepopulation  & $K$      & 50000\\
        cache size                   & $S$      & 80000\\
        block size                   & $B$      & 100\\
        \bottomrule
    \end{tabular}
    \label{table:hyperparameters}
\end{table}

\end{document}